\documentclass{article}
\usepackage{spconf,amsmath,graphicx}
\usepackage{booktabs}
\usepackage{cite}
\usepackage{subcaption}
\usepackage{multirow}
\usepackage{tipa}
\usepackage{bbm}
\usepackage{url}

\title{Joint Modeling of Accents and Acoustics\\for Multi-Accent Speech Recognition}
\makeatletter
\def\@name{ \emph{Xuesong Yang$^\star$, Kartik Audhkhasi$^\dagger$, Andrew Rosenberg$^\dagger$, Samuel Thomas$^\dagger$},  \\ \emph{Bhuvana Ramabhadran$^\dagger$, Mark Hasegawa-Johnson$^\star$} \vspace{3mm}}
\makeatother
\address{$^\star$University of Illinois at Urbana-Champaign, Urbana, IL\\
        $^\dagger$IBM T. J. Watson Research Center, Yorktown Heights, NY}
\begin{document}
\ninept
\maketitle
\begin{abstract}
 
The performance of automatic speech recognition systems degrades with increasing mismatch between the training and testing scenarios. Differences in speaker accents are a significant source of such mismatch. 
The traditional approach to deal
with multiple accents involves pooling data from several accents during training and building a single model in multi-task fashion, where tasks correspond to individual accents. 
In this paper, we explore
an alternate model where we jointly learn an accent classifier and
a multi-task acoustic model. Experiments on the American English Wall Street Journal and British English Cambridge corpora demonstrate that our
joint model outperforms the strong multi-task acoustic model baseline. We obtain a 5.94\% relative improvement  in word error rate on British English, and 9.47\% relative improvement on American English. This illustrates
that jointly modeling with accent information improves acoustic model 
performance.




\end{abstract}
\section{Introduction}
\label{sec:intro}

Recent breakthroughs in automatic speech recognition (ASR) have resulted in a word error rate (WER) on par with human transcribers~\cite{xiong2016achieving,saon2017english} on the English Switchboard benchmark. However, dealing with acoustic condition mismatch
between the training and testing data is a significant challenge that still remains unsolved. It is well-known that the performance of ASR systems
degrades significantly when presented with speech from speakers with different accents, dialects and speaking styles than those encountered during system training~\cite{huang2004accent}. In this paper, we specifically
focus on acoustic modeling for multi-accent ASR.

Dialects are defined as variations within a language that differ in geographical regions and social groups, which can be distinguished by traits of phonology, grammar, and vocabulary~\cite{holmes2013introduction}. Specifically, dialects may be associated with the residence, ethnicity, social class, and native language of speakers. For example, in British and American English, same words can have different spellings, like \emph{favour} and \emph{favor}; or different pronunciations, such as 
\emph{\textipa{"SEdju:l}} in UK English vs. 
\emph{\textipa{"skEdZUl}} in US English for the word \emph{schedule};
in Spanish, vocabulary may evolve differently between dialects, like for the phrase \emph{cell phone}, Castilian Spanish uses \emph{m\'ovil} while Latin American use \emph{celular}~\cite{soto2016selection}; in English, same phoneme may be realized differently, phoneme \emph{/\textipa{e}/} in \emph{dress} is pronounced as \emph{/\textipa{E}/} in England and \emph{/\textipa{e}/} in Wales; in Arabic, dialects may also differ in intonation and rhythm cues~\cite{biadsy2009using}. In this paper, we focus on the issue of differing pronunciations, while eschewing considerations of grammatical and vocabulary differences.

Acoustic modeling across multiple accents has been explored for many years, and various approaches can be summarized into three categories - \emph{Unified models}, \emph{Adaptive models}, and \emph{Ensemble models}. A unified model is trained on a limited number of accents, and can be generalized to any accent~\cite{elfeky2016towards,rao2017multi}. An adaptive model fine-tunes the unified model on accent-specific data assuming that the accent is known~\cite{huang2014multi,chen2015improving,yi2016ctc}. An ensemble model aggregates all accent-specific recognizers, and produces an optimal model by selection or combination for recognition~\cite{zheng2005accent,elfeky2015multi,soto2016selection}. Experiments have revealed that the unified model usually underperforms the adaptive model, which in turn underperforms the ensemble model~\cite{elfeky2016towards,rao2017multi}. 

We note that these prior approaches do not explicitly include accent information during training, but do so only indirectly, for example, through the different
target phoneme sets for various accents. This contrasts sharply with the way in which humans memorize the phonological and phonetic forms of accented speech: ``mental representations of phonological forms are extremely detailed,'' and include ``traces of individual voices or types of voices''~\cite{Pierrehumbert2016}.  
In this paper, we propose to link
the training of ASR acoustic models and accent identification models, in a manner
similar to the linking of these two learning processes in human speech perception. 
We show that this joint model
not only performs well on ASR, but also on accent identification when compared to
separately-trained models. 
Given the recent success in end-to-end 
models~\cite{graves2014towards,amodei2016deep,sak2015fast,miao2015eesen,hannun2014deep,maas2015lexicon,chorowski2014end,lu2015study,bahdanau2016end,chan2016listen,audhkhasi2017direct},
we use a bidirectional long short-term memory (BLSTM) recurrent neural network (RNN) acoustic model trained with the connectionist temporal classification (CTC) loss function for acoustic modeling. The accent identification (AID) network is also a BLSTM, but includes an average pooling layer to compute an utterance-level accent embedding. We also introduce a joint architecture where the lower layers of the network are trained using AID as the auxiliary task while multi-accent acoustic modeling remains the primary task of the network.

Next, we use the AID network as a hard switch between the accent-specific output layers of the CTC AM. Preliminary experiments on the Wall Street Journal American English and Cambridge British English corpora demonstrate that our joint model with the AID-based hard-switch achieves lower WER when compared with the state-of-the-art multi-task AM. We also show that the AID model also benefits from joint training.

The remainder of this paper is organized as follows: Section~\ref{sec:related_work} reviews relevant literature. Section~\ref{sec:method} introduces our AID model, multi-accent acoustic model, and switching strategy. Section \ref{sec:experiments} shows experiments and analysis, followed by the conclusion in Section~\ref{sec:conclusion}.

\section{Related Work}
\label{sec:related_work}

The most closely related work to ours is from~\cite{rao2017multi}, which illustrated that hierarchical grapheme-based AM with auxiliary phoneme-based AMs in four English dialects trained with CTC significantly outperformed accent-specific AMs and grapheme-based AM, respectively, while achieving competitive WER with phoneme-based multi-accent AM. Similarly, Yi et al~\cite{yi2016ctc} also trained a multi-accent phoneme-based AM with CTC loss, but instead, adapted accent-specific output layer using its target accent. 

Other relevant work compared the performance of training accent or dialect specific acoustic models and joint models.
These approaches predicted context-dependent (CD) triphone states using DNNs, and used a weighted finite state transducer (WFST)-based decoder.
For example, senones on accents of Chinese are predicted by assuming all accents within a language share a common CD state inventory~\cite{chen2015improving,huang2014multi}. Elfeky et al~\cite{elfeky2016towards} implemented a dialectal multi-task learning (DMTL) framework on three dialects of Arabic using the prediction of a unified set of CD states across all dialects prediction as the primary task and dialect identification as the secondary task. DMTL model deviated from ours in that it directly predicted CD states using convolutional-LSTM-DNNs (CLDNN), and was trained with either cross-entropy or state-level minimum Bayes risk, while ignoring the secondary dialect identification output at recognition time. This DMTL model was trained on all dialectal data and underperformed the dialect-specific model. Dialectal knowledge distilled (DKD) model was also designed in~\cite{elfeky2016towards}, which achieved results competitive to, but below, dialect-specific models.

The effectiveness of  dialect-specific models motivated investigations into how to use ensemble methods on multiple dialect-specific acoustic models for recognition. Soto et al~\cite{soto2016selection} explored approaches of selecting and combining the best decoded hypothesis from a pool of dialectal recognizers. This work is still different from ours in that we make our selection directly using predicted dialect. Huang et al~\cite{huang2004accent} used a similar strategy to ours by identifying accent first followed by acoustic model selection, however, this work only considered GMMs as the classifier. 

\section{Method}
\label{sec:method}

Our proposed system consists of multiple accent-specific acoustic models and accent identification model. We will describe these components and their joint model in this section. Acoustic model selection based on the hard-switch between accent-specific models is illustrated in Section \ref{sec:hardswitch}.

\subsection{Accent Identification}
\label{sec:aid}
Accurate identification of a speaker's accent is essential to the pipelined ASR systems, since accent identification (AID) errors can cause large mismatch to acoustic models. Given the hypothesis that accents can be discriminated by spectral features, researchers have attempted to model the spectral distribution of each accent using GMMs. Recently, DNNs have been explored as a much more expressive model compared to GMMs, especially in modeling probability distributions.

We implemented an independent AID that summarizes low-level acoustic features of an utterance by a stack of bidirectional LSTMs (BLSTMs) and DNN projection layers. An average-pooling layer is applied on top of transformed acoustic features, because the acoustic realization of a speaker's accent may not be observable in each frame. Applying average-pooling gives us a more robust estimate of accent-dependent acoustic features. We note that we assume that the speaker's accent is fixed over the entire utterance.

\begin{figure}[htbp]
    \centering
    \includegraphics[trim=0 90 100 30,clip,scale=0.4]{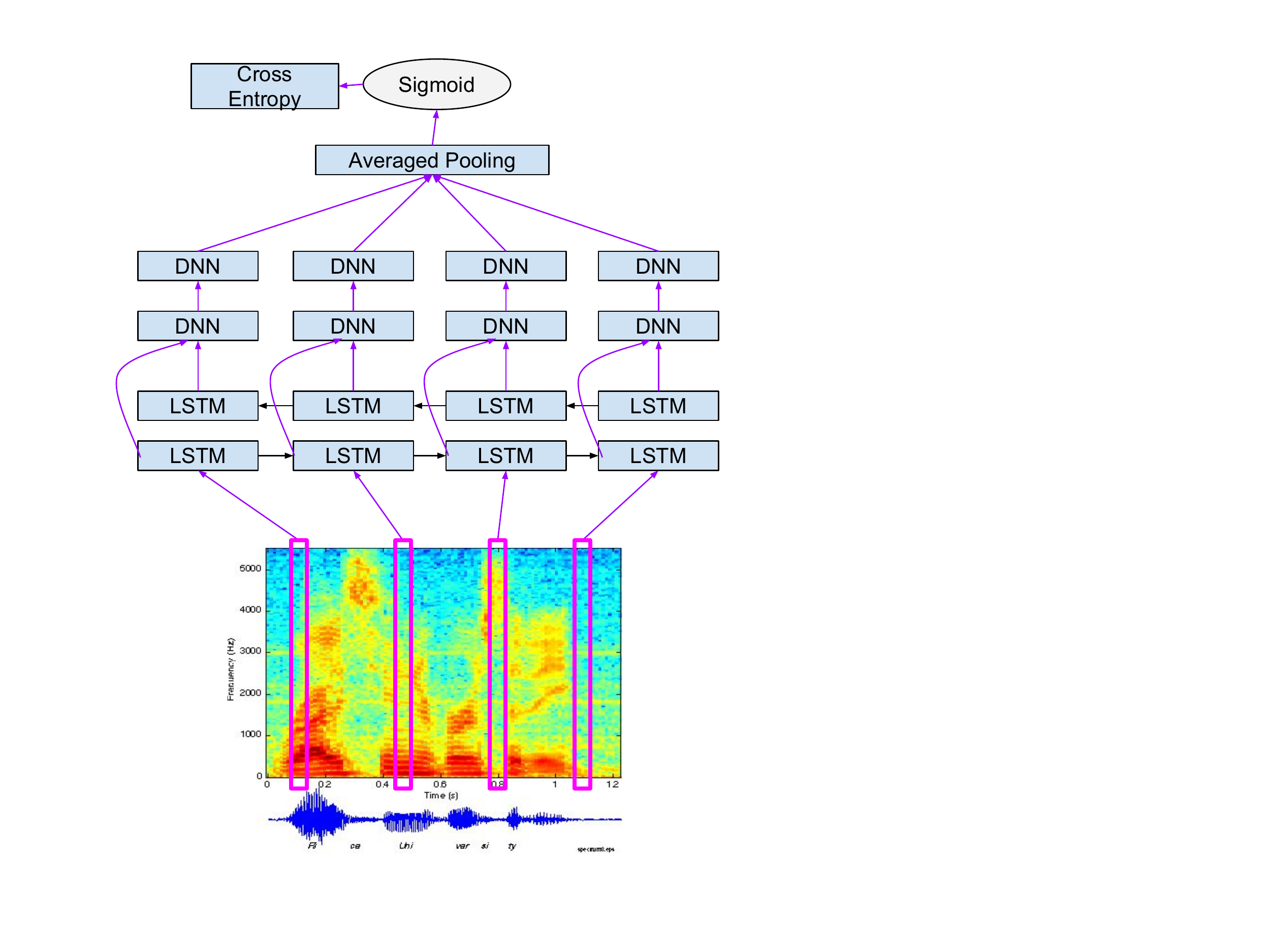}
    \caption{Proposed accent identification (AID) model with BLSTMs and average-pooling.}
    \label{fig:aid}
\end{figure}

Figure \ref{fig:aid} depicts details of this AID model. A single sigmoidal neuron is used at the output layer for classification because we are only classifying between accents of English - US and UK. We trained the AID network using the cross-entropy loss function.

\subsection{Multi-Accent Acoustic Modeling}
\label{sec:maam}
Recently, end-to-end (E2E) systems have achieved comparable performance to traditional pipelined systems such as hybrid DNN-HMM systems. These E2E systems come with the benefit of
avoiding time-consuming iterations between alignment and model building. RNNs using the CTC loss function are a popular approach to E2E systems~\cite{graves2014towards}. 
The CTC loss computes the total likelihood of the output label sequence given the input acoustics over all possible alignments. It achieves this by introducing a special \emph{blank} symbol that augments the label sequence to make its length equal to the length of the input sequence. Clearly, there are multiple such augmented sequences, and CTC uses the forward-backward algorithms to efficiently
sum the likelihoods of such sequences. The CTC loss is
\begin{equation}\label{eq:ctc}
	p(\mathbf{l} | \mathbf{x}) = \sum_{\pi \in \mathcal{B}^{-1}(\mathbf{l})} p(\pi | \mathbf{x})
\end{equation}
where $\mathbf{l}$ is the output label sequence, $\mathbf{x}$ is the input acoustic sequence, $\pi$ is a blank-augmented sequence for $\mathbf{l}$, and $\mathcal{B}^{-1}(\mathbf{l})$ is the set of all such sequences.
During decoding, the target label sequences can be obtained by either greedy search or a WFST-based decoder. 

\begin{figure}[htbp]
    \centering
    \includegraphics[trim=170 60 100 140,clip,scale=0.4]{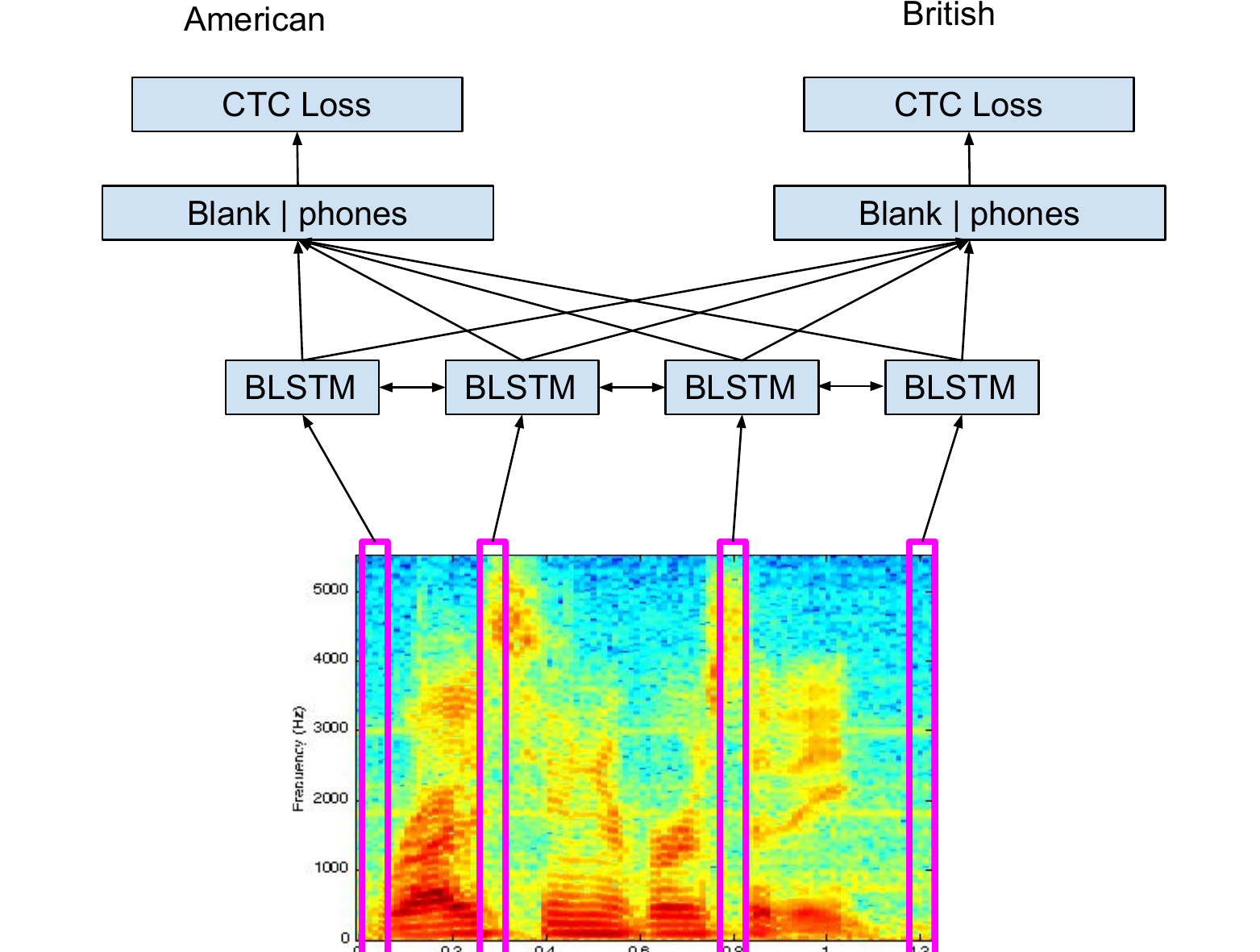}
    \caption{This figure shows the multi-accent acoustic model.}
    \label{fig:multidialect}
\end{figure}

Our multi-accent acoustic model combines two CTC-based AMs, one for each accent. We applied multiple BLSTM layers shared by two accents to capture accent-independent acoustic features, and placed separate DNNs for each AM to extract accent-specific features. Figure \ref{fig:multidialect} describes the structure of multi-accent acoustic model. Both AMs are jointly trained with an average of the two accent-specific CTC losses. 

At test time, this multi-accent model requires knowledge of the speaker's
accent to pick out of the two accent-specific targets. We experimented 
with both the oracle accent label, and using a trained AID network to make
this decision.

\subsection{Joint Acoustic Modeling with AID}
\label{sec:joint}
\begin{figure}[htbp]
	\centering
	\includegraphics[trim=170 35 100 180,clip,scale=0.4]{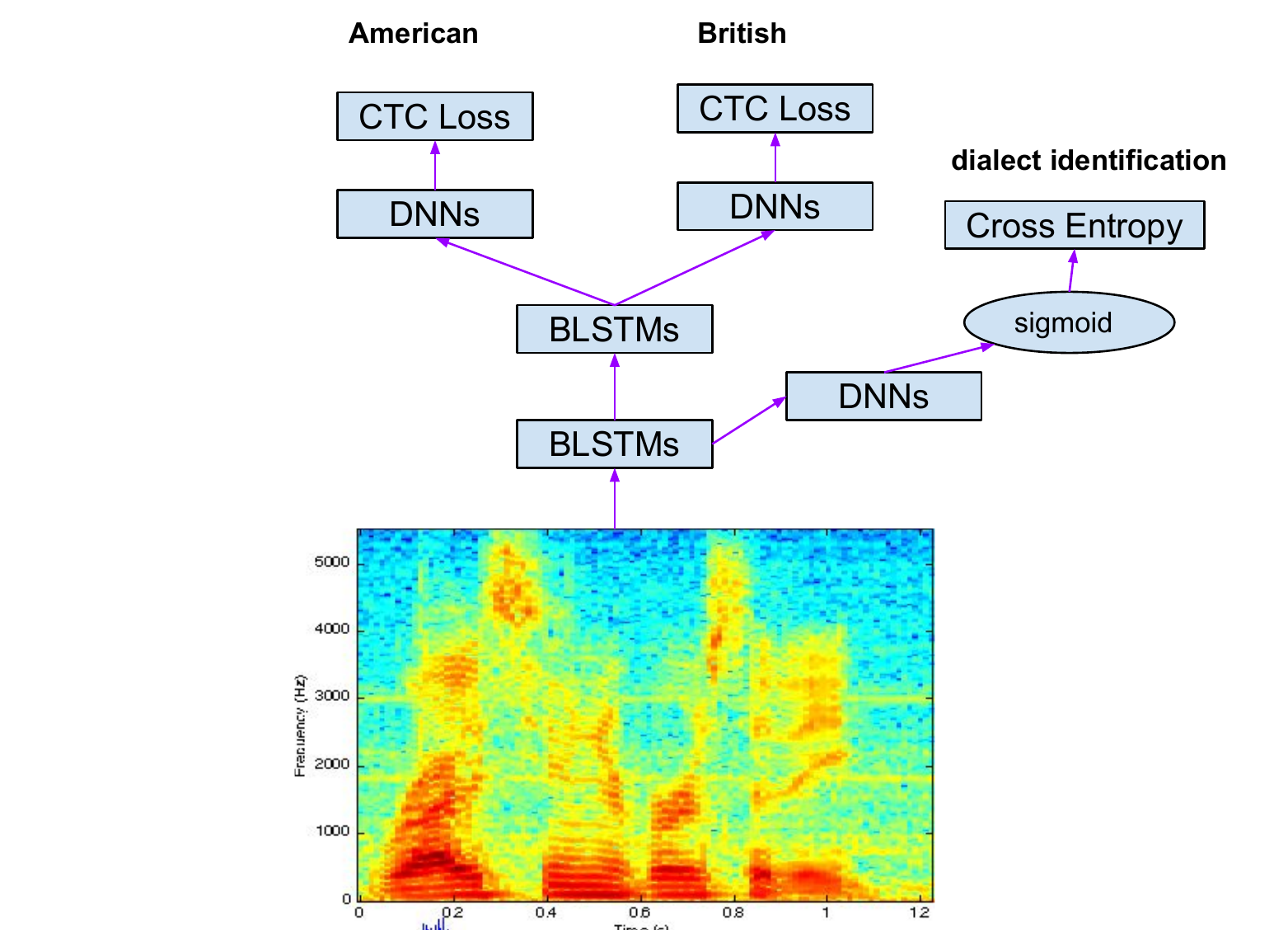}
    \caption{Proposed joint model for accent identification and acoustic modeling.}
    \label{fig:ham}
\end{figure}

The previous multi-accent model assumes that multi-tasking between the 
phone sets of the two accents is sufficient to make the network learn
accent-specific acoustic information. An alternate approach is to explicitly
supervise the network with accent information. This leads us to our joint model, with multi-accent acoustic modeling as primary tasks at higher layers, and with AID as an auxiliary task at lower level layers, as shown in Figure \ref{fig:ham}. 
This joint model aggregates two modules with the same structures to the forementioned models in Section \ref{sec:aid} and \ref{sec:maam}, and can be jointly trained in an end-to-end fashion with the objective function,
\begin{equation}
    \min_{\Theta} \mathcal{L}_\text{Joint}(\Theta) = (1 - \alpha) * \mathcal{L}_\text{AM}(\Theta) + \alpha * \mathcal{L}_\text{AID}(\Theta) \nonumber
\end{equation}
where $\alpha$ is an interpolation weight balancing between CTC loss of multi-accent AMs and the cross-entropy loss of AID, and $\Theta$ is the model parameters. CTC loss $\mathcal{L}_\text{AM}$ sums up the probabilities of all possible paths corresponding to Equation \eqref{eq:ctc}, while AID classification loss $\mathcal{L}_\text{AID}$ is cross-entropy. The two losses are at different scales, so the optimal value of $\alpha$ needs to be tuned on development data. 

\subsection{Model Selection by Hard-Switch}
\label{sec:hardswitch}

Given a trained CTC-based multi-accent acoustic model and AID classifier, we apply maximum likelihood estimation to switch between the accent-specific output layers $\mathbf{y}_\text{US}$ and $\mathbf{y}_\text{UK}$. Let $P_\text{AID}(\text{US}|\mathbf{x})$ denote the probability of the US accent estimated by AID. We threshold this probability at $0.5$ to obtain the accent hard-switch $s_\text{AID}(\text{US}|\mathbf{x})$. Hence, we pick the output layer as follows:
\begin{align}
    \mathbf{y} = \left\{\begin{matrix}
        \mathbf{y}_\text{US} & \text{if}\ s_\text{AID}(\text{US}|\mathbf{x}) = 1  \\ 
        \mathbf{y}_\text{UK} & else 
\end{matrix}\right. \nonumber
\end{align}

We note that this strategy applies to both the multi-accent model and the joint model.


\section{Experiments}
\label{sec:experiments}

We perform experiments on two dialects of English corpora--Wall Street Journal-1 American English and Cambridge British English. They contain overlapping, but distinct phone sets of 42 and 45 phones respectively. Both corpora contain approximately 15 hours each of audio.  We held-out 5\% of the training data as a development set. The window size of each speech frame is 25ms with a frame shift of 10ms. We extracted 40-dimensional log-Mel scale filter banks and performed per-utterance cepstral mean subtraction. We did not use any vocal tract length normalization. We then stacked neighboring frames and picked every alternate frame to get a 80-dimensional acoustic feature stream at half the frame rate. 
Various models are compared in terms of phone error rate (PER) and word error rate (WER). Particularly, we obtain the PER after simple frame-wise greedy decoding from the DNN projection outputs after removing repeated phones and the \emph{blank} symbol. The Attila toolkit~\cite{soltau2010ibm} is used to report WER by applying WFST-based decoding. Evaluation is performed on $\mathtt{eval93}$\footnote{\url{catalog.ldc.upenn.edu/ldc93s6a}}  American English and $\mathtt{si\_dt5b}$\footnote{\url{catalog.ldc.upenn.edu/LDC95S24}} British English.

Our joint model uses four BLSTM layers where the lowest layer is attached to the AID network and the highest single layer connects to two accent-specific softmax layers. A single DNN layer with 320 hidden units is used for each task. The weights for all models are initialized uniformly from $[-0.01, 0.01]$. $\mathtt{Adam}$~\cite{kingma2014adam} optimizer with initial learning rate $5e-4$ is used, and the gradients are clipped to the range $[-10, 10]$. We discard the training utterances that are longer than 2000 frames. New-bob annealing \cite{bourlNips1990} on the held-out data is used for early stopping, where the learning rate is cut in half whenever the held-out loss does not decrease. For the purpose of fair comparison, we used a four layer BLSTM for the baseline acoustic models as well. 

Various models are briefly described as follows:
\begin{itemize}
\item ASpec: phoneme-based accent-specific AMs that are trained separately on mono-accent data.
\item MTLP: phoneme-based multi-accent AMs that are jointly trained on two accents.
\item Joint: proposed phoneme-based joint acoustic model with AID.
\end{itemize}

\subsection{Empirical weights for balancing different losses}
\begin{figure}
	\centering
    \includegraphics[width=0.5\textwidth]{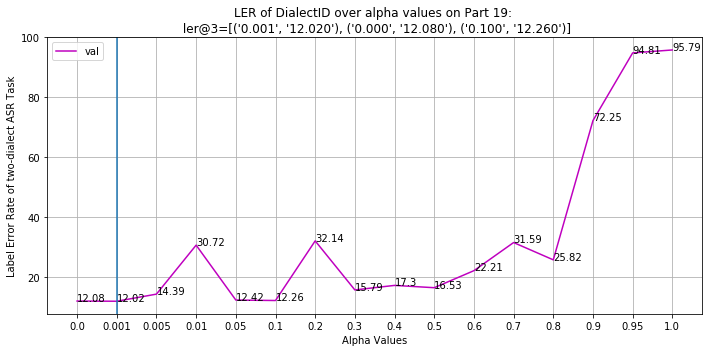}
    \caption{PER of joint acoustic model over AID loss weights $\alpha$}
    \label{fig:asr_ler_alpha}
\end{figure}

\begin{figure}
	\centering
    \includegraphics[width=0.5\textwidth]{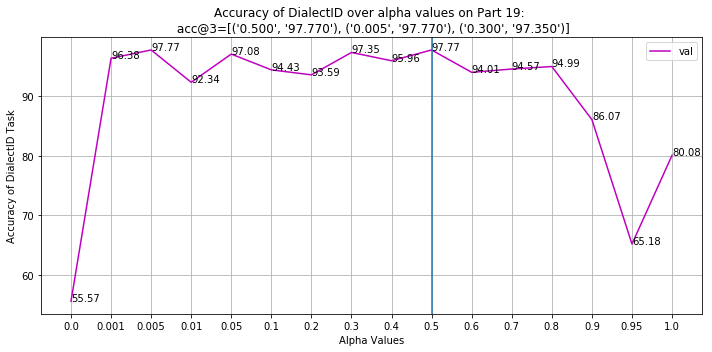}
    \caption{Accuracy of AID over various AID weights $\alpha$}
    \label{fig:aid_acc_alpha}
\end{figure}

Our joint model is sensitive to the interpolation weight $\alpha$ between the AM CTC and AID cross-entropy losses. We tuned $\alpha$ on development data. Figure \ref{fig:asr_ler_alpha}
depicts relationship between overall PER of two accents and different $\alpha$ values. When $\alpha$ goes larger, overall PER increases but with small fluctuations, especially at $\alpha$ of 0.01 and 0.2. The PER tends to be the largest if $\alpha$ is 1.0, which is expected since the weights of neural networks are updated only using the AID errors. We found the optimal value of $\alpha$ to be 0.001, which achieved minimum PER of 12.02\%. Figure \ref{fig:aid_acc_alpha} illustrates the trend of AID accuracy over different $\alpha$ values. Weights between 0.001 and 0.8 all perform well with accuracies greater than 92\%, while tail values lead to even worse performance. When $\alpha$ is 0.5 and 0.005, the best performance is achieved with 97.77\% accuracy. 

\subsection{Oracle performance for multi-accent acoustic models}
\label{subsec:oracle}

We first evaluate the oracle performance of various models in Table~\ref{tab:oracle}. These results assume that the correct accent of each utterance is provided for all models. In other words, the acoustic model corresponds to the correct accent, i.e.\ the relevant target accent-specific softmax layer is used. 
It can be seen that the proposed joint model significantly outperforms the accent-specific model (ASpec) by 17.97\% relative improvement in overall WER, and multi-task accent model (MTLP) by 6.81\%. This observation indicates that deep BLSTM layers shared with multiple accent AMs can learn expressive accent-independent features that refine accent-specific AMs. The auxiliary task, accent identification, also helps by introducing extra accent-specific information. The advantage of augmenting general acoustic features with specific information both implicitly learned by our joint model is observed in natural language processing~\cite{daume2009frustratingly} tasks as well. The value of implicit feature augmentation is a rich area for future investigation. 
\begin{table}[htbp]
	\centering
    \caption{Oracle performance in word error rates that assumes that the true accent ID is known in advance. Word error rates is calculated after decoding with a WFST-graph incorporating a LM; the relative improvement (\emph{rel.}) for each model over ASpec are reported in the parenthesis.}
    \label{tab:oracle}
    \vspace{-2mm}
    \begin{tabular}{|r|r|r|r|}
    	\hline
    	corpus & ASpec & MTLP (rel.) & Proposed Model (rel.)\\
        \hline
        British  & 11.5 & 10.1 (-12.17) & 9.5 (-17.39)\\
        American & 10.2 &  9.0 (-11.76)& 8.3 (-18.63)\\
        \hline
        average & 10.85 & 9.55 (-11.98) & 8.9 (-17.97)\\
        \hline
    \end{tabular}
\end{table}

\subsection{Hard-switch using distorted AID}
The oracle experiments in Section~\ref{subsec:oracle} demonstrate the value of our proposed joint model and the MTLP model when the AID classifier operates perfectly. This section demonstrates the impact of imperfect AID on the performance using hard-switch. Table \ref{tab:switch} shows the results. Given a well-trained independent AID (ind. AID), our joint model still significantly outperforms the two baseline models, and MTLP achieves better WER than ASpec. In comparison to oracle WERs of all models, British WERs are relatively constant without any distortion, however, American English WERs deteriorate accordingly. This is because independent AID has 100\% recall for British English utterances on the test data. 

It is interesting to note that the biggest improvement over \mbox{ASpec} in WER comes
when using the joint model (21.62\%) instead of the MTLP model (14.41\%) with an independent AID model. The improvement upon further using the AID from the joint model itself is still larger (22.52\%). This indicates that the joint model
has already learned sufficient accent-specific information through
the accent supervision in the lower layers.

\begin{table}[htbp]
    \centering
    \caption{WERs of hard-switch using distorted AID. The \emph{rel.} shows the relative improvement over ASpec; \emph{ind. AID} applies an independent neural AID trained separately. Our \emph{Proposed Model} applies the AID jointly learn with multi-accent AMs.}
    \label{tab:switch}
    \vspace{-2mm}
    \begin{tabular}{|r|c|cc|c|}
        \hline
        \multirow{2}{*}{Corpus} & \multicolumn{3}{c|}{Pipelines with ind. AID} & Proposed \\
        \cline{2-4}
        		   & ASpec &  MTLP (rel.) & Joint (rel.) &  Model (rel.)\\
        \hline 
          \multirow{2}{*}{British}  & \multirow{2}{*}{11.5} &  10.1& 9.5 & 9.5 \\
                   &       &  (-12.17) & (-17.39)& (-17.39)\\
          \hline
          \multirow{2}{*}{American} & \multirow{2}{*}{11.1} &  9.5   & 8.7   &  8.6   \\
                   &       &  (-14.41) & (-21.62) &  (-22.52)\\
        \hline
    \end{tabular}
\end{table}

\section{Conclusion}
\label{sec:conclusion}
This paper studies state-of-the-art approaches of acoustic modeling across multiple accents. We note that these prior approaches do not explicitly include accent information during training, but do so only indirectly, for example through the different phone inventories for various accents. We propose an end-to-end multi-accent acoustic modeling approach that can be jointly trained with accent identification. We use BLSTM-RNNs to design acoustic models that can be trained with CTC, and apply an average pooling to compute utterance-level accent embedding. Experiments show that both multi-accent acoustic models and accent identification benefit each other, and our joint model using hard-switch outperforms the state-of-the-art  multi-accent acoustic model baseline with a separately-trained AID network. We obtain a 5.94\% relative improvement in WER on British English, and 9.47\% on American English. 




\bibliographystyle{IEEEbib}
\bibliography{multi_dialect}

\end{document}